\documentclass{llncs}
\usepackage{latexsym}
\usepackage{amsmath,amsfonts}
\usepackage{multirow}
\usepackage{url,soul,dsfont}
\usepackage{graphics,amstext}
\usepackage[all]{xy}
\usepackage{graphicx,tikz,arydshln}

\newtheorem{defi}{Definition}

\title{Thematically Reinforced\\ Explicit Semantic Analysis}

\author{Yannis Haralambous (1)\quad Vitaly Klyuev (2)}
\institute{(1) Institut Mines-T\'el\'ecom - T\'el\'ecom Bretagne \& Lab-STICC UMR CNRS 6285\\\url{yannis.haralambous@telecom-bretagne.eu}\\
 (2) University of Aizu,
Aizu-Wakamatsu,
Fukushima-ken 965-8580, Japan\\\url{vkluev@u-aizu.ac.jp}}

\date{}

\begin{document}

\maketitle
\overfullrule10pt
\begin{abstract}
We present an extended, thematically reinforced version of Gabrilovich and Marko\-vitch's Explicit Semantic Analysis (ESA), where we obtain thematic information through the category structure of Wiki\-pedia. For this we first define a notion of categorical tfidf which measures the relevance of terms in categories. Using this measure as a weight we calculate a maximal spanning tree of the Wikipedia corpus considered as a directed graph of pages and categories. This tree provides us with a unique path of ``most related categories'' between each page and the top of the hierarchy. We reinforce tfidf of words in a page by aggregating it with categorical tfidfs of the nodes of these paths, and define a \emph{thematically reinforced ESA semantic relatedness measure} which is more robust than standard ESA and less sensitive to noise caused by out-of-context words. We apply our method to the French Wikipedia corpus, evaluate it through a text classification on a 37.5~MB corpus of 20 French newsgroups and obtain a precision increase of 9--10\% compared with standard ESA.
\end{abstract}

\clubpenalty10000
\widowpenalty10000

\fontsize{9.75}{11.75}\selectfont

\section{Introduction}

\subsection{Explicit Semantic Analysis}

Unlike semantic similarity measures, which are limited to ontological relations such as synonymy, hyponymy, meronymy, etc., \emph {semantic relatedness} measures detect and quantify semantic relations of a more general kind. The typical example is the one involving the concepts \textsc {car}, \textsc {vehicle} and \textsc {gasoline}. A car is a special kind of vehicle, so we have an hyperonym relation between the concepts, which can easily be quantified by a semantic similarity measure (for example, by taking the inverse of the length of the shortest path between the corresponding synsets in WordNet). But between \textsc {car} and \textsc {gasoline}, there is no semantic \emph{similarity}, since a car is a solid object and fuel is a liquid. Nevertheless, there is an obvious semantic relation between them since most cars use gasoline as their energy source, and such a relation can be quantified by a semantic \emph{relatedness} measure.

Gabrilovich \& Markovitch \cite {Gabrilovich:2007uo} introduce the semantic relatedness measure ESA (=~Explicit Semantic Analysis, as opposed to the classical method of Latent Semantic Analysis \cite {lsa-en}). ESA is based on the Wikipedia corpus. Here is the method: after cleaning and filtering Wikipedia pages (keeping only those with a sufficient amount of text and a given minimal number of incoming and outgoing links), they remove stop words, stem all words and calculate their tfidfs. Wikipedia pages can then be represented as vectors in the space of (nonempty, stemmed, distinct) words, the vector coordinates being normalized tfidf values. By the encyclopedic nature of Wikipedia, one can consider that every page corresponds to a concept. We thus have a matrix whose columns are concepts and whose lines are words. By transposing it we obtain a representation of words in the space of concepts. The ESA measure of two words is simply the cosine of their vectors in this space.

Roughly, two words are closely ESA-related \emph{if they appear frequently in the same Wikipedia pages} (so that their tfs are high), \emph{and rarely in the corpus as a whole} (for their dfs to be low). 

Despite the good results obtained by this method, it has given rise to some criticism. Thus, Haralambous \& Klyuev \cite {Haralambous:2011wm} note that ESA has poor performance when the relation between words is mainly ontological. As an example, in the English corpus, the word ``mile'' (length unit) does not appear in the page of the word ``kilometer'' and the latter appears only once in the page of the former: this is hardly sufficient to establish a nonzero semantic relatedness value; however, such a relation is obvious, since both words refer to units of length measurement. As pointed out in \cite {Haralambous:2011wm}, an ontological component, obtained from a WordNet-based measure, can, at least partially, fill this gap.

Another, more fundamental, criticism is that of Gottron et al. \cite {Gottron:2011im}, who argue that the choice of Wikipedia is irrelevant, and that any corpus of comparable size would give the same results. To prove it, they base ESA not on Wikipedia, but on the Reuters news corpus, and get even better results than with standard ESA. According to the authors, the semantic relatedness value depends only on the collocational frequency of the terms, and this whether documents correspond to concepts or not. In other words they deny the ``concept hypothesis,'' namely that ESA specifically uses the correspondence between concepts and Wikipedia pages. Also they state that while ``the application of ESA in a specific domain benefits from taking an index collection from the same topic domain while, on the other hand, a ``general topic corpus'' such as Wikipedia introduces noise,'' and this has precisely been our motivation for strengthening the thematic robustness of ESA. Indeed, in this article we will enhance ESA by adopting a different approach: the persistence of tfidfs of terms when leaving pages and entering the category graph.

\subsection{Wikipedia Categories}

A Wikipedia page can belong to one or more categories. Categories are represented by specific pages using the ``Category:'' prefix; these pages can again belong to other categories, so that we obtain a directed graph structure, the nodes of which can be standard pages (only outgoing edges) or categories (in- and outgoing edges). A page can belong to several categories and there is no ranking of their semantic relevance. For this reason, to be able to use categories, we first need an algorithm to determine the single semantically most relevant category, and for this we use, once again, ESA.

Wikipedia's category graph has been studied thoroughly in \cite {zesch} (for the English corpus).

\subsection {Related Work}\label{related}

Scholl et al. \cite {Scholl:2010tu} also enhance the performance of ESA using categories. They proceed as follows: let $ T $ be the matrix whose rows represent the Wikipedia pages and whose columns represent words. The value $ t_ {i, j} $ of cell $(i, j) $ is the normalized tfidf of the $ j $th word in the $ i $th page. For each word $ m $ there is therefore a vector $ \vec {v}_m $ whose dimension is equal to the number of pages. Now let $ C $ be the matrix whose columns are pages and whose lines are categories. The value of a cell $ c_ {i, j} $ is 1 when page $ j $ belongs to category $ i $ and 0 otherwise. They take the product of matrices $ \vec {v}_m \cdot C $ which provides a vector whose $ j $th component is $ \sum_ {i \mid D_i \in c_j} t_ {i, j} $, that is the sum of tfidfs of word $ m $ for all pages belonging to the $ j $th category. They use the concatenation of  vector $ \vec {v}_m $ and of the transpose of $ \vec {v}_m \cdot C $ to improve system performance on the text classification task. They call this method XESA (eXtended ESA).

We see that in this attempt, page tfidf is extended to categories by simply taking the sum of tfidfs of all pages belonging to a given category. This approach has a disadvantage when it comes to high-level categories: instead of being a way to find the words that  characterize a given category, the tfidf of a word tends to become nothing more than the average density of the word in the corpus, since for large categories, tf tends to be the total number of occurrences of the word in the corpus, while the denominator idf remains constant and equal to the number of documents containing the given word. Thus, this type of tfidf loses its power of discrimination for high-level categories. As we will see in Section~\ref{categorialtfidf}, we propose another extension of tfidf to categories, which we call \emph {categorical tfidf}. The difference lies in the denominator, where we take the number, not of all documents containing the term, but only of those \emph{not belonging} to the category. Thus our categorical tfidf (which is equal to the usual tfidf in the case of pages) is high when the term is common in the category and \emph {rare elsewhere} (as opposed to \emph {rare on the entire corpus} of Scholl et al.).

In \cite {Collin:2010}, the authors examine the problem of inconsistency of Wikipedia's category graph and propose a shortest path approach (based on the number of edges) between a page and the category ``\emph{Article},'' which is at the top of the hierarchy. The shortest path provides them with a semantic and thematic hierarchy and they calculate similarity as shortest length between vertices on these paths, a technique already used in WordNet \cite{wordnet}. However, as observed in \cite [p.~275] {wordnet}, the length (in number of edges) of the shortest path can vary randomly, depending on the density of  pages (synsets, in the case of WordNet) in a given domain of knowledge. On the other hand, the distance (in number of edges) between a leaf and the top of the hierarchy is often quite short, frequently requiring an arbitrary choice between paths of equal length.

What is common with our approach is the intention to simplify Wikipedia's category graph. But instead of counting edges, we weight the graph using ESA measure and use this weight, which is based on the statistical presence of words on pages belonging to a given category, to calculate a maximum spanning tree. The result of this operation is that any page (or category other than ``\emph{Article}'') has exactly one parent category that is semantically closest to it. This calculation is global, in the sense that the total weight of the tree is maximum.

We use this tree to define \emph {thematically reinforced ESA}. Our goal is to avoid words which, by accident, have a high tfidf in a given page despite the fact that they thematically do not really belong to it. This happens in the very frequent case where words have low frequencies (in the order of 1--3) so that the presence of an unsuitable word in a page results in a tfidf value as high (or even higher, if the word is seldom elsewhere) as the one of relevant words. Our hypothesis is that a word having an unduly high tfidf will disappear when we calculate its (categorical) tfidf in categories above the page, while, on the contrary, relevant words will be shared by other pages under the same category and their tfidfs will continue to be nonzero when switching to them. Such words will ``survive'' when we move away from leaves of the page-and-category tree and towards the root.

\section {Thematic Reinforcement}\label{arborification}

\subsection{Standard Tfidf, Concept Vector and ESA Measure}\label{standard}

Let us first formalize the standard ESA model.\footnote{All definitions in Section~\ref{standard} are from \cite {Gabrilovich:2007uo}.}

Let $\mathcal{W}$ be the Wikipedia corpus pruned by the standard ESA method, $p\in\mathcal{W}$ a Wikipedia page, and $ w\in p $ a word.\footnote{By ``word'' we mean an element of the set of character strings remaining after removing stopwords and stemming the Wikipedia corpus.} 
The tfidf $t_p(w)$ of the word $w$ on page $p$ is defined as:
$$ t_p (w) := (1 + \log (f_p (w))) \cdot \log \left (\frac {\# \mathcal {W}} {
\sum_{\substack{p\in\mathcal{W}\\w\in p}}1} \right) ,$$ 
where $ f_p (w) $ is the frequency of $ w $ on page $p $, $\# \mathcal {W} $ the cardinal of $\mathcal{W}$ and $\sum_{\substack{p\in\mathcal{W}\\w\in p}}1$, also known as the df (=~document frequency) of $w$, is the number of Wikipedia pages containing $ w $.

Consider the space $\mathbb{R}^{\#\mathcal W}$, where dimensions correspond to pages $p$ of $\mathcal{W}$. Then we define the ``concept vector'' $\vec{w}$ of word $w$ as
$$\vec{w}:=\sum_{p\in\mathcal{W}}t_p(w)\cdot {1}_p\in\mathbb{R}^{\#\mathcal W}$$
where ${1}_p$ is the unitary vector of $\mathbb{R}^{\#\mathcal W}$ corresponding to page $p$.

Let $w$ and $w'$ be words appearing in Wikipedia (and hence the Euclidean norms $\|\vec{w}\|$ and $\|\vec{w'}\|$ of their concept vectors are nonzero). The ESA semantic relatedness measure $\mu$ is defined as follows:
$$\mu(w,w'):=\frac{\langle \vec{w},\vec{w'}\rangle}{\|\vec{w}\|\cdot\|\vec{w'}\|}.$$

\subsection{Categorical Tfidf}\label{categorialtfidf}

Let $c$ be a Wikipedia category. We define $\mathcal{F}(c)$ as the set of all pages $p$ such that
\begin{itemize}
\item either $p$ belongs to $c$,
\item or $p$ belongs to $c_1$, and there a sequence of subcategory relations $c_1\to c_2\to\cdots\to c$, ending with $c$.
\end{itemize}

\begin{defi}Let $ w \in p $ be a word of $ p \in \mathcal {W} $, $ t_p (w) $ its standard tfidf in $p$, and $c$ a category of $\mathcal{W}$. We define the \emph {categorical tfidf} $ t_c (w) $ of $ w $ for category~$ c $ as follows:\label{categorialtfidf}
$$
t_c(w):=\left(1+\log\left(\sum_{p\in\mathcal{F}(c)}f_p(w)\right)\right)\cdot\left(\log\left(\frac{\#\mathcal{W}}{1+\sum_{\substack{p\in \mathcal{W}\setminus\mathcal{F}(c)\\w\in p}}1}\right)\right).
$$
\end{defi}

The difference with the tfidf defined by \cite {Scholl:2010tu} is in the calculation of df: instead of $ \sum_ {\substack{p \in \mathcal {W}\\w\in p}}  1$, that is the amount of pages containing $w$ in the entire Wikipedia corpus, we focus on those in $\mathcal {W} \setminus \mathcal {F} (c)$, namely the set difference between the whole corpus and pages that are ancestors of $c$ in the category graph, and we use $ 1 + \sum_ {\substack{p \in \mathcal {W} \setminus \mathcal {F} (c)\\w\in p}}  1$ instead (the unit is added to prevent a zero df in the case where the word does not appear outside $\mathcal{F}(c)$). We believe that this extension of tfidf to  categories improves discriminatory potential, even when the sets of pages become large (see discussion in Section~\ref{related}).

\subsection{Vectors of Pages and Categories}

Let $p\in\mathcal{W}$ be a page. We define the \emph{page vector} $\vec{p}$ as the normalized sum of concept vectors of its words, weighted by their tfidfs:
$$
\vec{d}:=\frac{\sum_{w\in p}t_p(w)\cdot \vec{w}}{\|\sum_{w\in p}t_p(w)\cdot \vec{w}\|}.
$$
Similarly let $c$ be a category of Wikipedia, we define the \emph{category vector} $\vec{c}$ as
$$
\vec{c}:=\frac{\sum_{w\in \mathcal{F}(c)}t_c(w)\cdot \vec{w}}{\|\sum_{w\in \mathcal{F}(c)}t_c(w)\cdot \vec{w}\|}.
$$
where $w\in\mathcal{F}(c)$ means that there exists a page $p$ such that $p\in\mathcal{F}(c)$ and $w\in p$.

\subsection{Wikipedia Arborification}

\begin{defi}\label{wsr}
Let $p$ be a Wikipedia page and $c,c'$ Wikipedia categories. Let $p\to c$ be the membership of page $d$ to category $c$, and $c\to c'$ the subcategory relation between $c$ and $c$. We define
the \emph {weight of semantic relatedness} of these relations as
\begin{align*}
p(p\to c)&=\langle \vec{p},\vec{c}\rangle.\\
p(c\to c')&=\langle \vec{c},\vec{c'}\rangle,
\end{align*}
where $ \langle \kern1pt. \kern1pt, \kern1pt. \kern1pt \rangle $ is the Euclidean scalar product of two vectors.  
\end{defi}

This product is equal to the cosine metric since the vectors are all unitary. By this property we also have $ \mathrm {Im} (p) \subset [0,1] $.

The relations considered in Definition~\ref{wsr} correspond to vertices of the Wiki\-pe\-dia category graph. Let $\mathcal{W}'$ be the weighted Wikipedia digraph (whose vertices are pages and categories, whose edges are memberships of pages and inclusions of categories, and whose weight is the weight of semantic relatedness).

At this point we can already reinforce the standard tfidf of words on pages, by the categorical tfidf of the same words in related categories. But how can we choose these categories? Taking all those containing a page would result in cacophony since categories can be more or less relevant and sometimes have no semantic relation whatsoever. Not to mention the fact that the Wikipedia category graph is quite complex, and using it as such would be computationally prohibiting.

The solution we present to this problem is to simplify $\mathcal{W}'$ by extracting a maximal spanning tree. It should be noted that standard minimal/maximal spanning tree algorithms such as Kruskal or Prim cannot be applied because $\mathcal{W}'$ is directed, has a global sink, namely the ``Article'' page, and we want the orientation of the directed spanning tree to be compatible with the one of the directed graph\footnote{It is a known fact that every rooted tree has exactly two possible orientations: one going from the root to the leaves and one in the opposite direction.}.

To obtain the maximal spanning tree, we utilized Chu-Liu \& Edmonds' algorithm \cite [p.~113-119]{chuliu}, published for the first time in 1965. This semi-linear algorithm returns a minimum weight forest of rooted trees covering the digraph. The orientation of these rooted trees is compatible with the one of the graph. In the general case, connectivity is not guaranteed (even though the graph may be connected).
But in the case of a digraph containing a global sink, the forest becomes a single tree, and we get a true \emph{directed maximal spanning tree} of the graph.
If our case, the global sink is obviously the category that is hierarchically at the top, namely ``\emph{Article}.''\footnote{It should be noted, however, that the path between a page and the root on the maximal spanning tree is not a maximal path per se, since the importance is given to the global maximality of weight, for the whole tree. If our goal were to find the most appropriate taxonomy for a specific page, i.e., the most relevant path from this page to the top, then it would be more appropriate to use a shortest/longest path algorithm, such as Dijk\-stra. This has already been proposed in \cite {Collin:2010}, but for the metric of the number of edges; in our case we would rather use the measure given by the weight of the graph.}

Let $\mathcal{T}$ be the maximal spanning tree of $\mathcal{W}'$ obtained by our method. As in any tree, there is a unique path between any two nodes. In particular, there is a unique path between any page-node and the root; we call it the \emph{sequence of ancestors} of the page.

\subsection{Thematically Reinforced ESA}

We will use the page ancestors in the maximal spanning tree to update tfidf values of words in the page vectors. Indeed, a word in a given page may have a high tfidf value simply because it occurred one or two times, this does not guarantee a significant semantic proximity between the word and the page. But if the word appears also in ancestor categories (and hence, in other pages belonging to the same category), then we have stronger chances for semantic pertinence.

\begin{defi}\label{trtfidf} Let $p$ be a Wikipedia page, $w$ a word $w\in p$, $t_p(w)$ the standard tfidf of $w$ in $p$, $(\pi^i(p))_i$ the sequence of ancestors of $p$, and $(\lambda_i)_i$ a decreasing sequence of positive real numbers converging to~$0$. We define the \emph{thematically reinforced tfidf} $t_{p,\lambda_*}(w)$ as
$$
t_{p,\lambda_*}(w)=t_p(w)+\sum_{i\geq 0}\lambda_i t_{\pi^i(p)}(w).
$$
\end{defi}

The sum is finite because the Wikipedia maximal spanning tree is finite and hence there is a maximal distance from the root, after which the $\pi^i$ become vacuous.

\begin{defi} With the notations of Definition~\ref{trtfidf}, we define the \emph{thematically reinforced concept vector} $\vec{w}_{\lambda_*}$ as
$$\vec{w}_{\lambda_*}:=\sum_{p\in\mathcal{W}}t_{p,\lambda_*}(w)\cdot {1}_p\in\mathbb{R}^{\#\mathcal W}.$$
In other words, it is the usual concept vector definition, but using thematically reinforced tfidf.
\end{defi}

With these tools we can define our extended version of ESA, as follows:

\begin{defi} With the notations of Definition~\ref{trtfidf} and $w,w'\in\mathcal{W}$, we define the
\emph{thematically reinforced ESA semantic relatedness measure} $\mu_{\lambda_*}$ as:
$$\mu_{\lambda_*}(w,w'):=\frac{\langle \vec{w}_{\lambda_*},\vec{w}'_{\lambda_*}\rangle}{\|\vec{w}_{\lambda_*}\|\cdot\|\vec{w}'_{\lambda_*}\|}.$$
In other words, it is the usual ESA measure definition, but using thematically reinforced concept vectors and tfidf.
\end{defi}

\section {Corpus}

We have chosen to work on the French Wikipedia corpus (version of December 31, 2011), which is smaller than the English one and, to our knowledge, has not yet been used for ESA. To adapt ESA to French Wikipedia, we followed the same steps as \cite {Gabrilovich:2007uo} and \cite {calli} except for one: we have preceded stemming by lemmatization, to avoid loss of information due to poor stemming of inflected words. (In English, inflection is negligible, so that stemming can be performed directly.)

Originally, the authors of \cite {Gabrilovich:2007uo} pruned the 2005 English Wikipedia corpus down to 132,689 pages. In our case, by limiting the minimum size of pages to 125 (nonstop, lemmatized, stemmed and distinct) words, 15 incoming and 15 outgoing links, we obtained a number of Wikipedia pages comparable to that of the original ESA implementation, namely 128,701 pages (out of 2,782,242 in total) containing 1,446,559 distinct words (only 339,679 of which appear more than three times in the corpus).

\begin{figure*}[t]
\centering\resizebox{\textwidth}{!}{\includegraphics{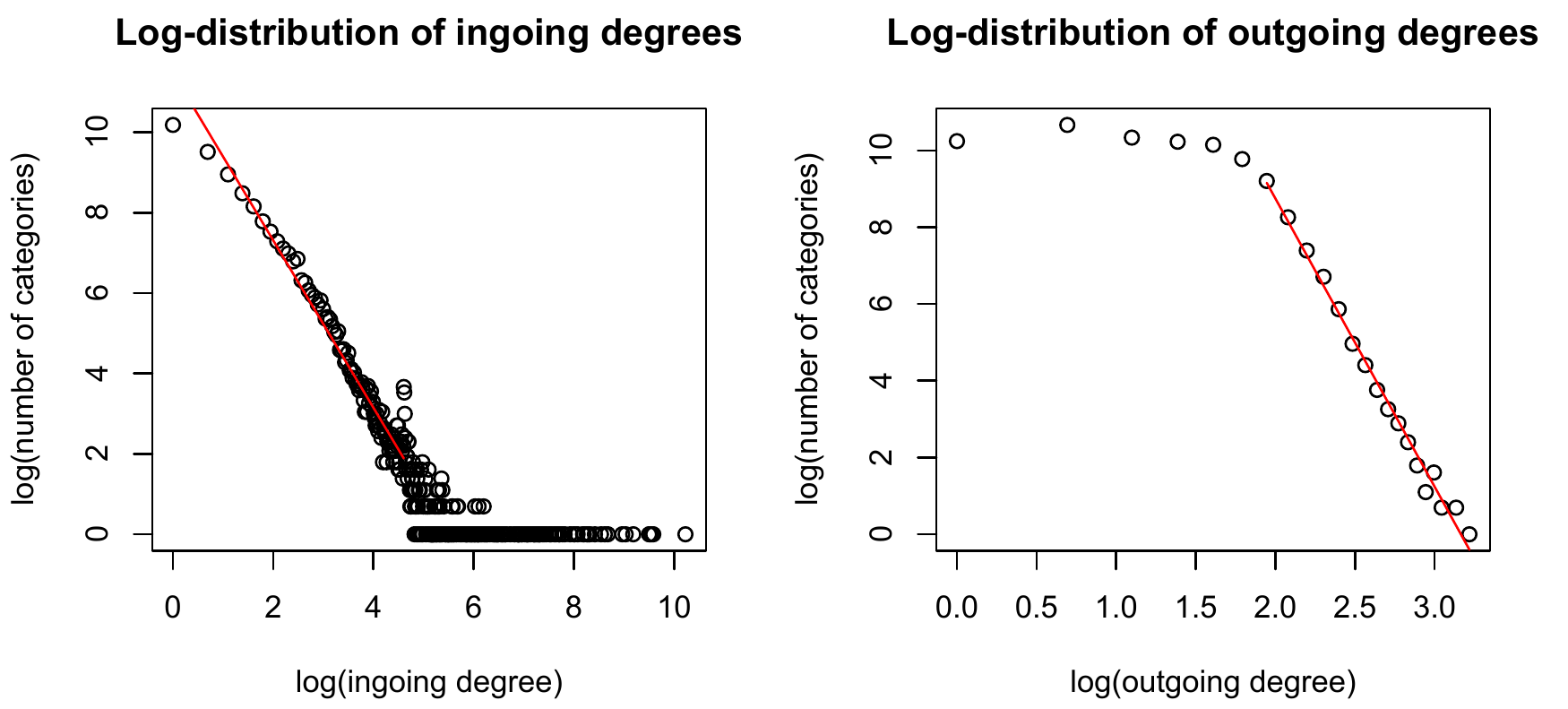}}
\caption{Ingoing and outgoing degree distribution of French Wikipedia categories. \label {fig0}}

\end {figure*}

Furthermore, the French corpus contains 293,244 categories, 680,912 edges between categories and 12,935,688 edges between pages and categories. As can be seen on Fig.~\ref{fig0}, by the logarithmic distribution of incoming and outgoing degrees, this graph follows a power distribution $ p ^ {- \alpha} $ with $ \alpha =  2.08$ for incoming degrees and $ \alpha =  7.51$ for outgoing degrees. According to \cite[p.~248]{newman}, the former value is typical, while the latter can be considered very high, and this was another motivation for simplifying the Wikipedia graph by extracting the maximal spanning tree, instead of performing heavy calculations on the entire graph.

The French Wikipedia category graph is fairly complex and, in particular, contains cycles. Indeed, according to \cite {Medelyan:2009ts}, ``cycles are not encouraged but may be tolerated in rare cases.'' The very simple example of categories ``\emph{Zoologie}'' (=~Zoology) and ``\emph{Animal}'' (in French Wikipedia) pointing to each other, shows that the semantic relation underlying subcategories is not always hyperonymy. Here \textsc {animal} is the object of study of the discipline \textsc {zoology}. We attempted the following experiment: starting from the 2,782,242 (unfiltered) French Wikipedia pages, we followed random paths formed by the category links. The choice of each subsequent category was made at random, but did not change during the experiment. 78\% of these paths contained cycles, but it turned out that it was always the same 50 cycles, 12 of which were of length~3 (triangles) and all others of length~2 (categories pointing to each other, as in the example above, which was detected by this method). Hence, we were able to turn this directed graph acyclic by merely removing 50 edges.

\section{Evaluation}

Gabrilovich and Markovitch \cite {Gabrilovich:2007uo} evaluate their method on WS-353, a set of 352 English word pairs, the semantic relatedness of which has been evaluated by 15--16 human judges. Their criterion is the Spearman correlation coefficient between the rank of pairs obtained by ESA and that obtained by taking the average of human judgments. Our first attempt was to translate these pairs into French, but the result was rather disappointing.\footnote{Indeed, some twenty words are untranslatable into a simple term (the current version of ESA covers only single-word terms), such as ``seafood'' which can be translated only as ``\emph{fruits de mer}.'' Furthermore there are ambiguities of translation resulting from word polysemy: When we translate the pair ``flight/car'' by ``\emph{vol}/\emph{voiture},'' we obtain a high semantic relatedness due to the criminal sense of ``\emph{vol}'' (=~theft) while the sense of the English word ``flight'' is mainly confined to the domain of aviation. Finally, some obvious collocations disappear when translating word for word, such as ``soap/opera'' which is unfortunately not comparable to ``\emph{savon}/\emph{op\'era}''\ldots}

We have therefore chosen to evaluate our implementation of ESA in a more traditional way, by performing a text classification task. We have extracted a total of 20,000 French language messages from the 20 most popular French newsgroups. The characteristics of our evaluation corpus can be seen on Table~\ref{corpus-tab}, where the second column represents the number of messages for a given newsgroup, the third the number of words, and the fourth, the number of distinct stemmed nonstop words that also occur in Wikipedia.

\begin{table}[th]
\caption{Characteristics of the evaluation corpus\label{corpus-tab}}
\small\centering

 \begin {tabular} {llrrr} 
Theme&Newsgroup & \# mess. & \# words & \# terms\\\hline
Medicine&fr.bio.medecine&1,000&738,258&14.785\\
Writing&fr.lettres.ecriture&1,000&688,849&14,948\\
French language&fr.lettres.langue.francaise&1,000&594,143&14,956\\
Animals&fr.rec.animaux&1,000&391,270&10,726\\
Classical music&fr.rec.arts.musique.classique&1,000&379,794&15,056\\
Rock music&fr.rec.arts.musique.rock&1,000&318,434&12,764\\
Do-it-yourself&fr.rec.bricolage&1,000&358,220&8,349\\
Movies&fr.rec.cinema.discussion&1,000&680,480&18,284\\
Gardening&fr.rec.jardinage&1,000&495,465&12,042\\
Photography&fr.rec.photo&1,000&415,767&10,931\\
Diving&fr.rec.plongee&1,000&485,059&11,326\\
Soccer&fr.rec.sport.football&1,000&612,842&13,548\\
Astronomy&fr.sci.astronomie&1,000&444,576&10,781\\
Physics&fr.sci.physique&1,000&598,079&13,916\\
Economics&fr.soc.economie&1,000&737,795&14,797\\
Environment&fr.soc.environnement&1,000&683,806&15,756\\
Feminism&fr.soc.feminisme&1,000&612,844&16,716\\
History&fr.soc.histoire&1,000&675,957&16,458\\
Religion&fr.soc.religion&1,000&763,477&16,124\\
Sects&fr.soc.sectes&1,000&738,327&16,732\\\hline
Global&&20,000&11,413,442&67,902
\end{tabular}
\end{table}

To perform text classification we need to extend the definitions of tfidf and document vector to the evaluation corpus. Let $\mathcal{C}$ be the evaluation corpus and $d$ a document $d\in\mathcal{C}$. We define the tfidf $t_d(w)$ of a word $w\in d$ in $\mathcal{C}$ as
$$
t_d(w):=(1+\log(f_d(w)))\cdot\mathrm{log}\left(\frac{\#\mathcal{C}}{\mathrm{df}(w)}\right),
$$
where $f_d$ is the frequency of $w$ in $d$; $\#\mathcal{C}$ the total number of documents; $\mathrm{df}(w)$ the number of documents in $\mathcal{C}$, containing $w$.

Furthermore, our ESA implementation provides us with a concept vector $\vec{w}$ for every word $w$. We define the \emph{document vector} $\vec{d}$ as:
$$
\vec {d}: = \frac {\sum_ {w \in d} t_d (w) \cdot \vec {w}} {\|\sum_ {w \in d} t_d (w) \cdot \vec {w}\|}.
$$
where the denominator is used for normalization.

Using these vectors, text classification becomes standard classification in $\mathbb{R}^{\#\mathcal{W}}$ for the cosine metric. We applied the linear multi-class SVM classifier\label {svm} SVM$^{\text{multiclass}}$ \cite{Joachims:1999} to the set of these vectors and the corresponding document classes, and after a tenfold cross-validation, we obtained an average precision of 65.58\% for a $C$ coefficient of $3.0$. The classification required $324$ support vectors. Admittedly the precision obtained is rather low, which is partly due to the thematic proximity of some classes (like, for example, Religion and Sects, or Writing and French language). However, our goal is not to compare ESA to other classification methods, but to show that our approach improves ESA. So, this result is our starting point and we intend to improve it.

\long\def\tetrecitw#1{\resizebox{\columnwidth}{\height}{#1}}

\begin{table}[t]
\caption{Evaluation results (ordered by decreasing precision)\label{res}}
\begin{center}
\footnotesize
\tetrecitw{\begin{tabular}{cccccccl}
$\lambda_1$&$\lambda_2$&$\lambda_3$&$\lambda_4$&$\lambda_5$&C&\# SVs&Precision\\\hline
1.5&0&0.5&0.25&0.125&3.0&786&\textbf{75.015\%}\\%ESSAI13
1&0&0.5&0.25&0.125&3.0&709&74.978\%\\%ESSAI10
1.5&1&0.5&0.25&0.125&3.0&827&74.899\%\\%ESSAI14
0.25&1.5&0.5&0.25&0.125&3.0&761&74.87\%\\%ESSAI12
0.5&0&0.5&0.25&0.125&3.0&698&74.867\%\\%ESSAI18
1&0.5&0.25&0.125&0.0625&3.0&736&74.845\%\\%ESSAI7
0.5&1&0.5&0.25&0.125&3.0&736&74.795\%\\%ESSAI6
1&1.5&0.5&0.25&0.125&3.0&865&74.791\%\\%ESSAI9
0.5&0.5&0.5&0.25&0.125&3.0&682&74.789\%\\%ESSAI17
0.5&1.5&0.5&0.25&0.125&3.0&778&74.814\%\\%ESSAI20
1.5&0.5&0.2&0.1&0.05&3.0&775&74.780\%\\%ESSAI8
\end{tabular}~~\begin{tabular}{cccccccl}
$\lambda_1$&$\lambda_2$&$\lambda_3$&$\lambda_4$&$\lambda_5$&C&\# SVs&Precision\\\hline
0&1&0.5&0.25&0.125&3.0&710&74.716\%\\%ESSAI11
2&1&0.5&0.25&0.125&3.0&899&74.705\%\\%ESSAI21
2&0&0.5&0.25&0.125&3.0&852&74.675\%\\%ESSAI23
0.5&0.25&0.125&0.0625&0.0312&3.0&653&74.67\%\\  %ESSAI3
2&0.5&0.5&0.25&0.125&3.0&899&74.641\%\\%ESSAI22
0.25&0.125&0.0625&0.0312&0.015&3.0&615&74.613\%\\  %ESSAI5
1&1&1&0.5&0.25&3.0&796&74.61\%\\   %ESSAI2
0&1.5&1&0.5&0.25&3.0&792&74.548\%\\%ESSAI16
1.5&1.5&1&0.75&0.25&3.0&900&74.471\%\\%ESSAI15
2&1.5&1&0.5&0.25&3.0&\textbf{995}&74.36\%\\ %ESSAI1
0&0&0&0&0&3.0&324&65.58\%
\end{tabular}}
\end{center}
\end{table}

\begin{figure}[ht]
\resizebox{.5\textwidth}{!}{\includegraphics{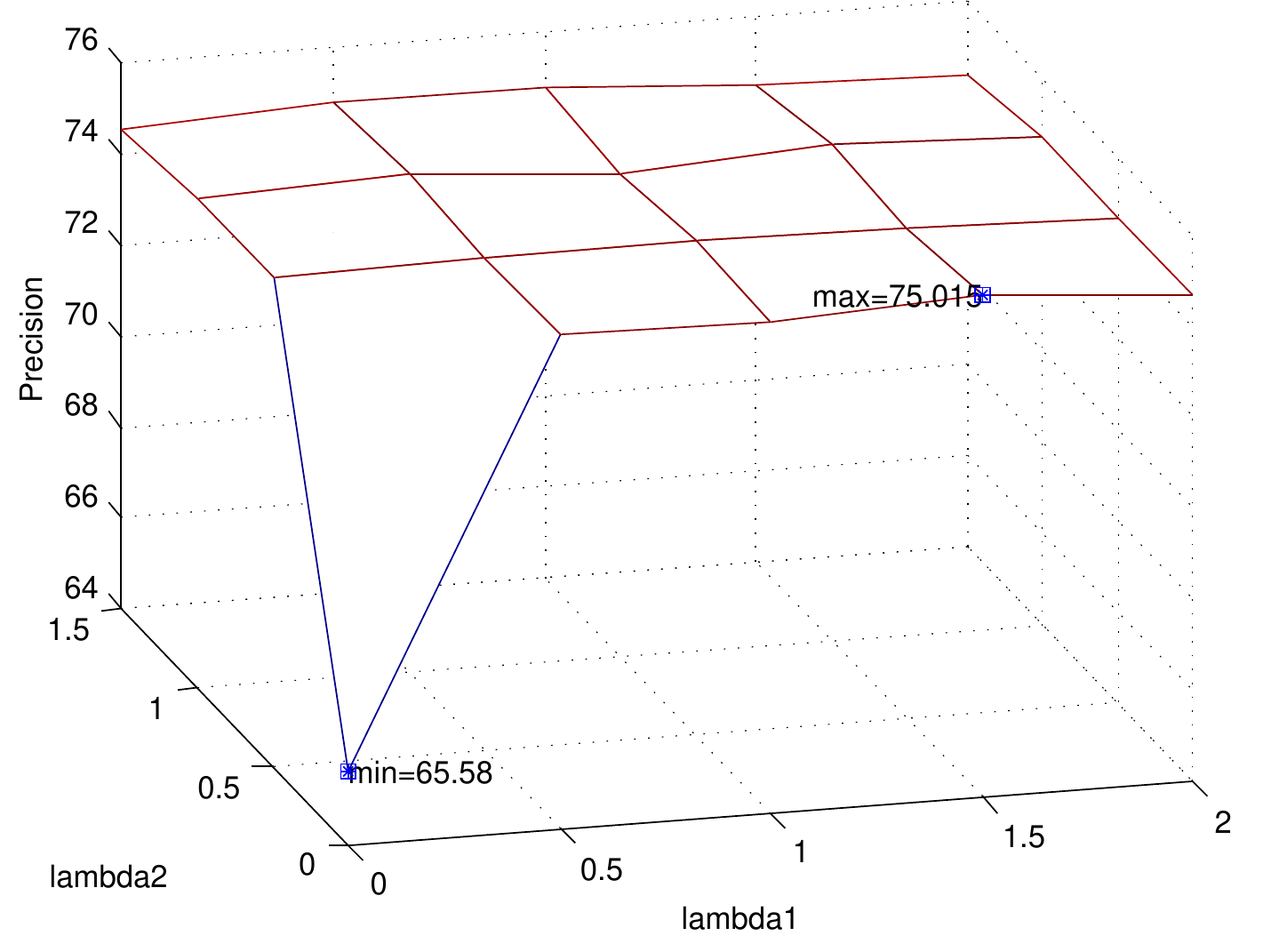}}\resizebox{.5\textwidth}{!}{\includegraphics{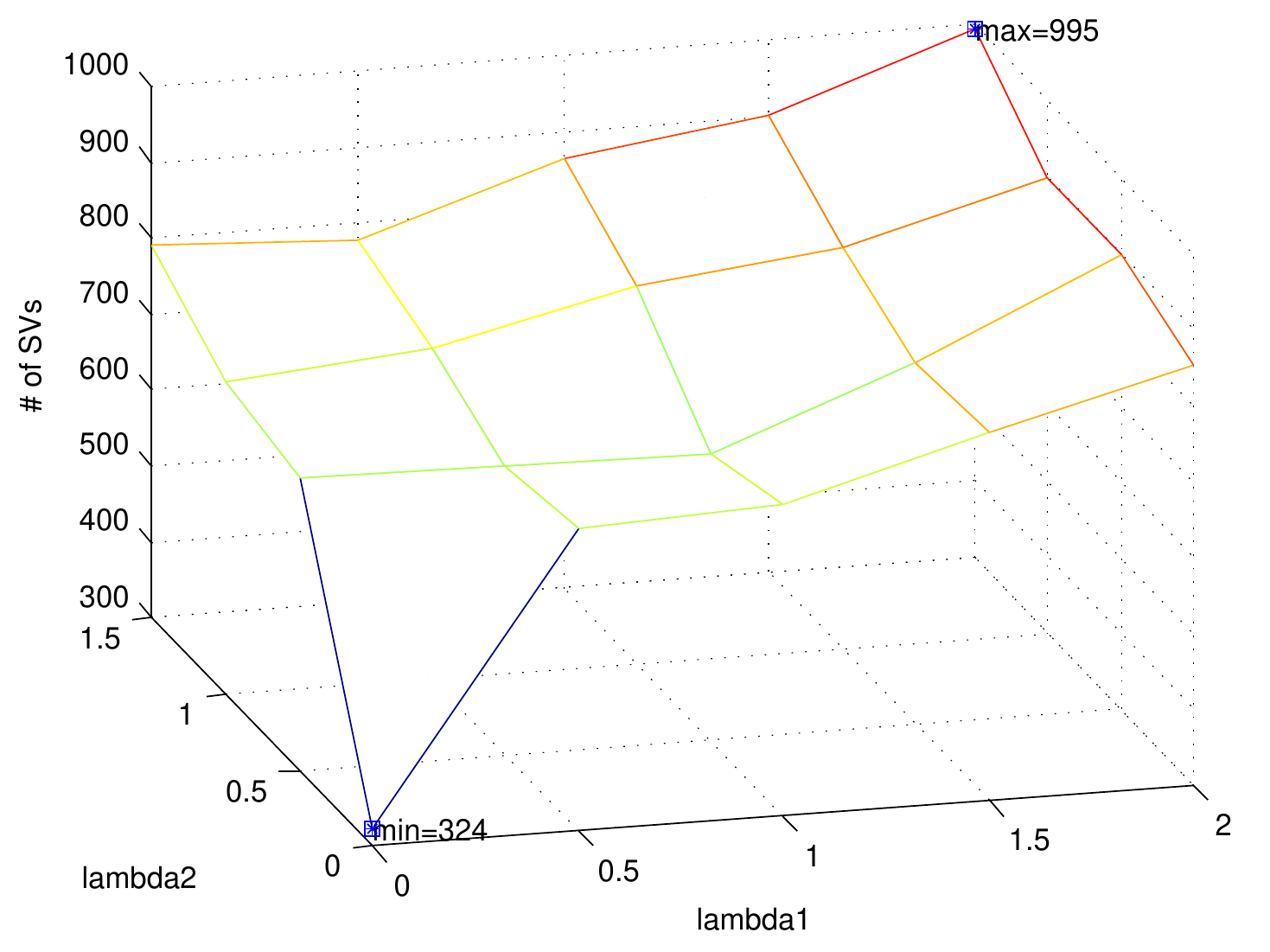}}

\caption{Precision (to the left) and number of support vectors used (to the right), as functions of the parameters $\lambda_1$ and $\lambda_2$.\label{matlab}}
\end{figure}

We followed the same modus operandi using thematically reinforced methods and obtained the results displayed on Table~\ref{res}. The results show a significant improvement over the standard ESA version (that corresponds to $\lambda_i=0$ for all $i$. This confirms our approach. On Fig.~\ref{matlab} the reader can see the precision obtained as function of the two first parameters $\lambda_1$ and $\lambda_2$, as well the number of support vectors used. We notice that the precision varies slightly (between 74.36\% and 75.015\%, that is less than 1\%) as long as $\lambda_1$ or $\lambda_2$ are nonzero, and abruptly goes down to 65.58\% when they are both zero. For nonzero values of $\lambda_i$ the variation of precision follows no recognizable pattern. On the other hand, the number of support vectors shows a pattern: it is clearly correlated with $\lambda_1$ and $\lambda_2$, the highest value being 995, number of support vectors used when both $\lambda_1$ and $\lambda_2$ take their highest values. Since CPU time is roughly proportional to the number of support vectors, it is most interesting to take small (but nonzero) values of $\lambda_i$ so that, at the same time, precision is high and the number of support vectors (and hence CPU time) is kept small.

\section {Conclusion and Hints for Further Research}

By reinforcing the thematic context of words in Wikipedia pages, context obtained through the category structure, we claim to be able to improve the performance of the ESA measure.

We evaluated our method on a text classification task based on messages from the 20 most popular French language newsgroups: thematic reinforcement allowed us to improve the classification precision by 9--10\%.

Here are some hints for research to be done:
\begin {enumerate}
\item propose the notion of the ``most relevant category'' to Wikipedia users and use their feedback to improve the system;
\item when we take the ``most relevant category'' for each page, we don't consider by how much it is better than the others. For small differences of semantic relevance weight between categories one could imagine alternative ``slightly worse'' spanning trees and compare the results;
\item by comparing relevance between alternative ``most relevant'' categories for the same page one could quantify a ``global potential'' of the Wikipedia corpus. Compare with Wikipedia corpora in other languages;
\item aggregate the thematically reinforced measure with collocational and ontological components, as in \cite {Haralambous:2011wm};
\item define another measure, based on links between pages (or categories), proportional to the number of links (or link paths) between pages and inversely proportional to the length of these paths. Compare it to ESA (which uses the number of links between pages to filter Wikipedia, but does not include it in semantic relatedness calculations) and thematically reinforced ESA;
\item and, more generally, explore the applications of graph theory to the formi\-dable mathematical-linguistic objects represented by the different graphs extracted from Wikipedia.
\end {enumerate}

\bibliographystyle{splncs}
\bibliography{paper2}

\end{document}